\documentclass[preprint,12pt]{elsarticle}
\usepackage{lineno,hyperref}
\modulolinenumbers[5]
\usepackage{makecell}
\usepackage{graphics}
\usepackage{amsmath}
\usepackage{adjustbox}
\journal{Computers \& Security (Elsevier)}
\usepackage{multirow}
\usepackage{graphicx}
\usepackage{enumitem}
\usepackage{float}
\usepackage{wrapfig}
\usepackage{multirow}
\usepackage{blindtext}
\usepackage{tabularx}
\usepackage{tabulary}
\usepackage[normalem]{ulem}
\usepackage{amsmath}
\usepackage{ragged2e}
\usepackage{inputenc}
\usepackage{booktabs}
\usepackage{rotating}
\usepackage{tikz}
\usepackage{adjustbox}
\usepackage{booktabs} 
\usepackage[caption=false]{subfig}
\usepackage[T1]{fontenc}
\usepackage{color, colortbl}
\usepackage[misc,geometry]{ifsym}
\usepackage{dirtytalk}
\definecolor{grey}{gray}{0.9}
\usepackage[graphicx]{realboxes}
\usepackage{siunitx}
\usepackage{rotfloat}
\usepackage{wrapfig}

\begin{document}

\begin{frontmatter}

\title{Real or Virtual: A Video Conferencing Background Manipulation-Detection System}

\author[1,2]{Ehsan Nowroozi}
\ead{ehsan.nowroozi@sabanciuniv.edu}
\author[4]{Yassine Mekdad}
\ead{ymekdad@fiu.edu}
\author[3]{Mauro Conti}
\ead{conti@math.unipd.it}
\author[5]{Simone Milani}
\ead{simone.milani@dei.unipd.it}
\author[4]{\\A. Selcuk Uluagac}
\ead{suluagac@fiu.edu}
\author[1,2]{Berrin Yanikoglu}
\ead{berrin@sabanciuniv.edu}

\address[1]{Faculty of Engineering and Natural Sciences, Sabanci University, Istanbul, Turkey}
\address[2]{Center of Excellence in Data Analytics, Sabanci University, Istanbul, Turkey}
\address[3]{Department of Mathematics, University of Padua, Padua, Italy}
\address[4]{Cyber-Physical Systems Security Lab, Florida International University, Miami, FL, USA}
\address[5]{Department of Information Engineering, University of Padua, Padua, Italy}

\begin{abstract}
Recently, the popularity and wide use of the last-generation video conferencing technologies created an exponential growth in its market size. Such technology allows participants in different geographic regions to have a virtual face-to-face meeting. Additionally, it enables users to employ a virtual background to conceal their own environment due to privacy concerns or to reduce distractions, particularly in professional settings. Nevertheless, in scenarios where the users should not hide their actual locations, they may mislead other participants by claiming their virtual background as a real one.
Therefore, it is crucial to develop tools and strategies to detect the authenticity of the considered virtual background. 

In this paper, we present a detection strategy to distinguish between real and virtual video conferencing user backgrounds. We demonstrate that our detector is robust against two attack scenarios. The first scenario considers the case where the detector is unaware about the attacks (i.e., the forensically-edited frames are not part of the training set). In the second scenario, we make the detector aware of the adversarial attacks, which we refer to Adversarial Multimedia Forensics (i.e, the forensically-edited frames are included in the training set). Given the lack of publicly available dataset of virtual and real backgrounds for video conferencing, we created our own dataset and made them publicly available~\cite{DatasetCalls}. Then, we demonstrate the robustness of our detector against different adversarial attacks that the adversary considers. Ultimately, our detector's performance is significant against the CRSPAM1372~\cite{Ehsan_IWBF2018} features, and post-processing operations such as geometric transformations with different quality factors that the attacker may choose. Moreover, our performance results shows that we can perfectly identify a real from a virtual background with an accuracy of 99.80\%.
\end{abstract}


\end{frontmatter}



\section{Introduction}
\label{sec.intro}
Nowadays, people's lives are made easier by various video conferencing applications (e.g., Zoom, Google-Meet, Microsoft Teams), especially during the pandemic. Smart working, online meetings, and virtual conferencing platforms have - and continue to be - widely adopted solutions to connect people, while preserving social distancing. On the other hand, the wide spread use and popularity of such solutions have also brought the need for privacy-preserving and security measures against several attacks such as JPEG compression and various post-processing operations intended to forge images~\cite{WIFS2020_CoMat}. In video conferencing platforms, replacing a real background with a virtual one enables the users to conceal their real location, remove distractions, or hide other people present in the environment. However, a worker might also want to fool his/her employer by changing the real background image or consider a video with a virtual background~\cite{Zatko15}. This possibility also paves the way for creating forged videos with people pretending to be in different locations. Therefore, developing robust detectors that identify real from virtual background is an interesting emerging, yet a challenging task in the world of multimedia authentication. To that end, there is a need to develop such a tool that can distinguish a real background from a virtual background.\\ 

In the past decade, the recent development of machine and deep learning approaches has significantly enhanced the detection of forged videos~\cite{Nowroozi2021AForensics}. However, the identification and the computation of robust video features that enable flawless detection is still a challenging research issue. In particular, the attacker can fool the detector by faking his virtual background that seems to be a real background. The technique proposed in~\cite{Nataraj2019DetectingGG} achieves a considerably high detection rate by creating color co-occurrence matrices of the presented background as input to the Convolutional Neural Network (CNN). The authors obtained outstanding performance for a variety of manipulation operations by training the CNN using co-occurrence matrices generated from the input image, instead of the co-occurrence matrices that are extracted from noise residuals. The authors considered two different Generative Adversarial Network (GAN) image datasets that are based on unpaired image-to-image translations, the styleGAN~\cite{Zhu2017unpaired} and the starGAN~\cite{choi2018stargan}. In multimedia forensics, residual images are frequently used in calculating the co-occurrence matrices to identify or localize the detection~\cite{Ehsan_IWBF2018,HigherOrder2017}; and these approaches often consider the SPAM features \cite{Jessica_SPAM2010}, or the CSRMQ \cite{Rich_Goljan2014} features, which were initially suggested for steganalysis. In \cite{LI2020107616}, the color components are used to generate co-occurrence matrices from high-pass filtering residuals, as well as for each truncated residual image. Then, the co-occurrences are obtained by combining the color channels into a feature to train the Support Vector Machine (SVM).\\ 

The solutions in~\cite{WIFS2020_CoMat, VIPPrint} achieve outstanding results by feeding six co-occurrence matrices (hereafter, we referred to as \textit{six co-mat}) into CNN from spatial and spectral bands. Additionaly, they are applied on highly fake image quality that is produced by the GAN. This approach computes co-occurrence matrices by considering cross-band co-occurrences (spectral) and co-occurrences (spatial) on the grey-level computed separately on the single bands. Although targeting ProGAN \cite{ProGAN} and StyleGAN \cite{StyleGAN} datasets mainly, the authors argue that it can be employed in any multimedia forensics detection task. Compared to the three co-occurrence matrices approach that has spatial features~\cite{Nataraj2019DetectingGG}, the six co-occurrence matrices approach with both spatial and spectral features significantly enhance the resilience against various types of post-processing and laundering attacks~\cite{WIFS2020_CoMat}.\\

Inspired by the later approach, we are motivated to understand the usability of such method on videoconferencing systems and other related applications against robust adversarial attacks. More specifically, we investigate the security implications of videoconferencing systems and build a detector tool that can distinguish a real background from video with a virtual background. To classify the virtual from real background, our proposed method searches for inconsistencies in co-occurrence matrices frame-by-frame. Consequently, the aware detector achieved high performance for revealing JPEG compressed frames compared to the existing approaches so far.\\ 

Furthermore, we perform additional experiments to verify the robustness of our proposed methods against several common attacks focusing on Macbook-Pro camera. For this, we also captured different videos in other cameras to test the detector. Based on the results that we achieved in MSI GF65 10UX laptop (low quality) and Apple IPAD Pro (high quality), we find that if the main component of a camera fingerprint are close to each other, our detector is robust in terms of detection. Additionally, we found that if we want the detector to operate against different cameras (video conferencing), a fine-tuning of the detector is needed or another methodology should be used for detecting different camera model identification based on sensor pattern noise~\cite{filler2008using}. The experimental results show that our aware model detector performs well over a wide variety of attack scenarios that we considered in this study~\cite{DatasetCalls}.

\subsection{Research aim and questions\\}

Recently, a significant research effort has been dedicated to developing forensic tools to retrieve data and detect possible tampered multimedia documents~\cite{uluagac2013passive}. Furthermore, several counter-forensic approaches have been developed to provide an appropriate analysis~\cite{bohme2013counter,nowroozi2021survey,roy2020digital}. These approaches are often very effective given the vulnerability of existing multimedia forensics tools, which are commonly not designed to operate under adversarial settings. In this scenario, it is necessary to develop forensic techniques qualified for achieving high performance under adversarial settings, referred to as \textit{adversarial multimedia forensics}~\cite{nowroozi2020machine,Nowroozi2021AForensics}. This turns out to be a challenging task due to the weakness of the traces usually handled by the forensic analysis.\\ 

The goal of this research is to provide a robust detector (trained with videos captured in a different software and cameras) in adversarial multimedia forensics, in the unaware and aware scenarios, when the detector is unaware of the attacks (we refer to as \textit{unaware scenario}) and in other scenarios when the detector is aware of the attacks (we refer to as \textit{aware scenario})~\cite{nowroozi2020machine,Nowroozi2021AForensics}.\\ 

Our work is generally focused on adversarial multimedia forensic~\cite{Nowroozi2021AForensics,Barni2018AdversarialAhead,nowroozi2020machine}, and not on camera model identification~\cite{filler2008using,Tuama2017CameraNetworks,Bondi2017FirstNetworks}. As a result, the following questions are motivating us to write this paper:
\begin{itemize}
    \item \textit{RQ1: How are the most current techniques efficient to identify the malicious videos from real ones?}
    \item \textit{RQ2: How can a user detect a real background from a virtual one in videoconferencing software under adversarial settings?}
    \item \textit{RQ3: What is the performance of such a detector against different attack/manipulation scenarios?}
\end{itemize}

\subsection{Problem Definition\\}

In a day-to-day professional life, videoconferencing meetings have become ubiquitous, especially during situations when physical meetings are not possible. Recently, several companies have allowed their employees to work remotely and cooperate globally through videoconferencing software. Such benefits enable the digital workforce and promote collaboration. However, video conferencing software might pose privacy threats related to disclosing personal data. For instance, the visual content transmitted during a video conferencing call can directly be shared from the private environment of the users to third parties, thus exposing sensitive information. To that end, video conferencing software solves the privacy leakage by replacing the observable environment of a videoconferencing call with a virtual background that covers the sensitive living environment. Nevertheless, this solution is inefficient as it might leak some parts of the real background in video frames. In~\cite{Hilgefort2021SpyingCalls}, the authors explored the exploitation of these leaks in virtual backgrounds to reconstruct parts of the real background such as Zoom videoconferencing software. The authors showed that for given high-quality camera, the tiny leakage in the real background frames is superficial if the subject moves slowly. On the other hand, some users can mislead other participants by claiming their virtual background as a real one. In this context, and to the best of our knowledge, the existing research on the robustification of such software against adversarial attacks is limited. In our work, we construct a detector that can distinguish a given video with a real background from a video with a virtual background under adversarial settings, and thus by considering two specific scenarios: 
\begin{itemize}
\item The case where the detector is unaware of the adversarial attacks (unaware scenario). 
    \item The case where the detector is aware of the adversarial attacks (aware scenario).
    
\end{itemize}

Additionally, we consider novel attacks that a potential malicious actor can perform to fool videoconferencing users. Furthermore, we test the detector with match case (i.e., Zoom software using Macbook-Pro) and mis-match case (i.e., Google-Meet and Microsoft Teams using both Apple IPAD Pro and MSI GF65 10UX laptop) with two different cameras.

\subsection{Contributions\\} 

The following is a list of the paper's contributions:
\begin{itemize}
        \item We investigate the most current techniques used for identifying malicious images from real ones and CRSPAM1372~\cite{WIFS2020_CoMat, Ehsan_IWBF2018}, and we use these approaches to identify real videos (real background) from virtual ones when users utilize the virtual background.
    \item We demonstrate that our novel approach can perfectly work as a detector by proposing an experimental evaluation of the classifier that can identify the type of background used in the video conferencing software.
    \item We evaluate the detector's performance with several cameras and with different video conferencing software under various types of attacks (e.g., changing the background of the user, different lighting conditions, considering different post-processing operations), and also when the detector is aware of these novel attack scenarios.
\end{itemize}

\subsection{Organization\\} 
This paper is organized as follows: In Section~\ref{background}, we present the state of art on the security and privacy of videoconferencing backgrounds. Afterwards, we provide the proposed system and threat models with the feature extraction methods in Section~\ref{system}. Section \ref{sec.exMethodd} describes the methodology used in our research, while Section \ref{sec.ExpAna} reports and discusses the findings of the experiments. Section \ref{sec.con} concludes the research with some remarks and future works.

\section{State of The Art}
\label{background}
In this section, we provide the state of art of backgrounds used by several videoconferencing software, then we discuss the related works on the security of virtual backgrounds.

\subsection{Videoconferencing Backgrounds\\}
In the following, we provide a description about different videoconferencing backgrounds that can be used by users during a videoconferencing call.
\begin{itemize}
    \item Real Background: It consists of the background captured by the camera of the user during a videoconferencing meeting. This background reflect the real image of the user's background and has not been changed, altered or edited by any additional software.
    \item Virtual Background: One of the techniques used by several videoconferencing software is to allow the user to hide or change the real background with a particular image as a background. This image is known as the \textit{virtual background}, and aims to cover the real background for different purposes (e.g., privacy, avoiding distraction during the meeting, etc.).
\end{itemize}

Although replacing the real background with a virtual one has several benefits, it triggers security and privacy concerns for the user. For instance, an adversary can potentially reveal or recover hidden parts of the user's background~\cite{Hilgefort2021SpyingCalls}. Another scenario include the use of the virtual background as real one. In this case, there is a need to develop an efficient tool that can detect a virtual background from a real one.

\subsection{Related work\\}
Detecting the authenticity of a virtual background whether it's real or fake can be referred to detecting a given image if it is GAN-generated or not~\cite{Nataraj2019DetectingMatrices,WIFS2020_CoMat}. Prior works in multimedia forensics propose several methods to identify GAN-generated images from natural ones. It is worth mentioning that the most current methods are based on Convolutional Neural Networks, and achieve good detection accuracy. For instance, the approach developed in~\cite{Nataraj2019DetectingGG} compute from three color image bands the co-occurrence matrices and fed them to the CNN model. Another approach proposed in~\cite{WIFS2020_CoMat} demonstrated that although Generative adversarial networks can produce high quality images with likely invisible traces, it is challenging to reconstruct  a consistent relationship among the color bands. As an extension of the method proposed in~\cite{Nataraj2019DetectingGG}, the authors improved the detection of images generated through GANs, and thus by considering as input the cross-band co-occurrences and the gray-level co-occurrences to the CNN detector. These co-occurrences are separately computed on the single bands.
Consequently, the proposed CNN detector achieves a good detection accuracy, and robustness for the intra-band method against post-processing. However, these approaches applied for GAN content detection are not robust against different attacks and post-processing~\cite{Nataraj2019DetectingGG,WIFS2020_CoMat}. Additional research efforts have been considered to investigate the privacy leakage of virtual backgrounds in videoconferencing software~\cite{Hilgefort2021SpyingCalls}. The authors demonstrated the possibility of partial reconstruction of the visual environment despite of the virtual background. They evaluated their adversarial attack against several videoconferencing software (e.g., Zoom, Webex, and Google Meet), and the possibility of identifying objects in the user's virtual background. Another work conducted investigated the privacy issues of the virtual background in videoconferencing platforms~\cite{he2021privacy}. However, these works cannot perfectly identify a real from a virtual background under adversarial attacks~\cite{Hilgefort2021SpyingCalls,he2021privacy}.\\ 

In this line of research, we observe that if the videos are captured with a high-quality camera such as MacBook-Pro and the subject moves slowly, we do not have tiny leakage in the real background parts. Consequently, the work proposed in~\cite{Hilgefort2021SpyingCalls} is not efficient in this case. Moreover, the reconstruction of the real background will be a challenging task. On the other hand, and differently from the underlining approach, we are interested in the security of Machine Learning and Deep Learning in~\cite{Hilgefort2021SpyingCalls}, where the adversary can fool the detector easily through different kinds of attacks. Additionally, we noticed similar privacy concerns about the virtual background, where the users could conceal their location with the virtual background and cheat other participants in the meeting, such as considering the real background as a virtual one, or different lighting conditions.

\section{System and Threat Model}
\label{system}
In this section, we describe our proposed system and threat model considered to detect real background from a virtual one in videoconferencing videos. Then, we present the feature extraction methods namely CRSPAM1372 and \textit{six co-mat}. 

\subsection{System Model\\}
\begin{figure}[h!]%
	\centering
	\subfloat[The detector is trained with CRSPAM1372 features]{{\includegraphics[width=0.75\textwidth]{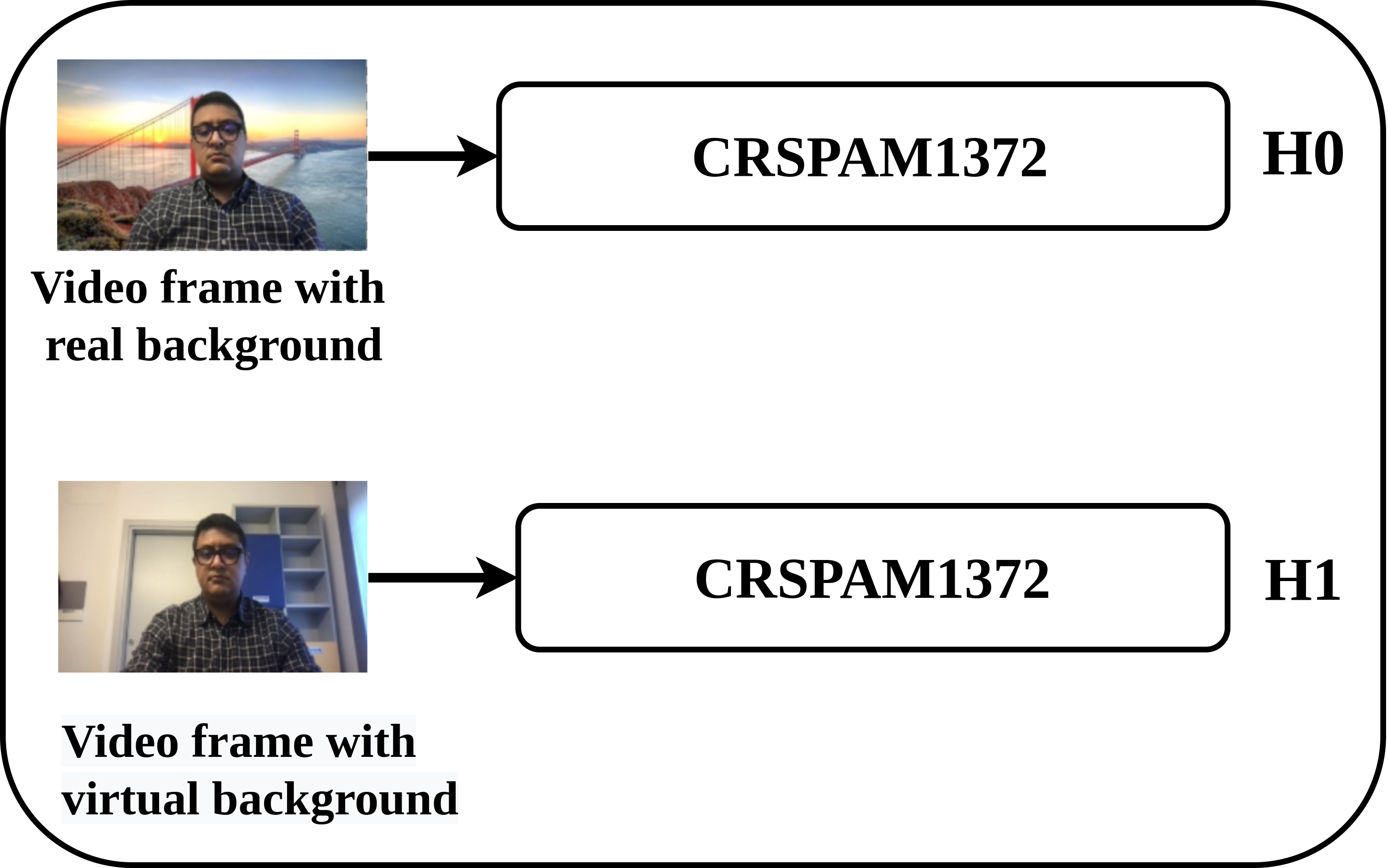} }}%
	\qquad
	\subfloat[The detector is trained with six-co mat features]{{\includegraphics[width=0.75\textwidth]{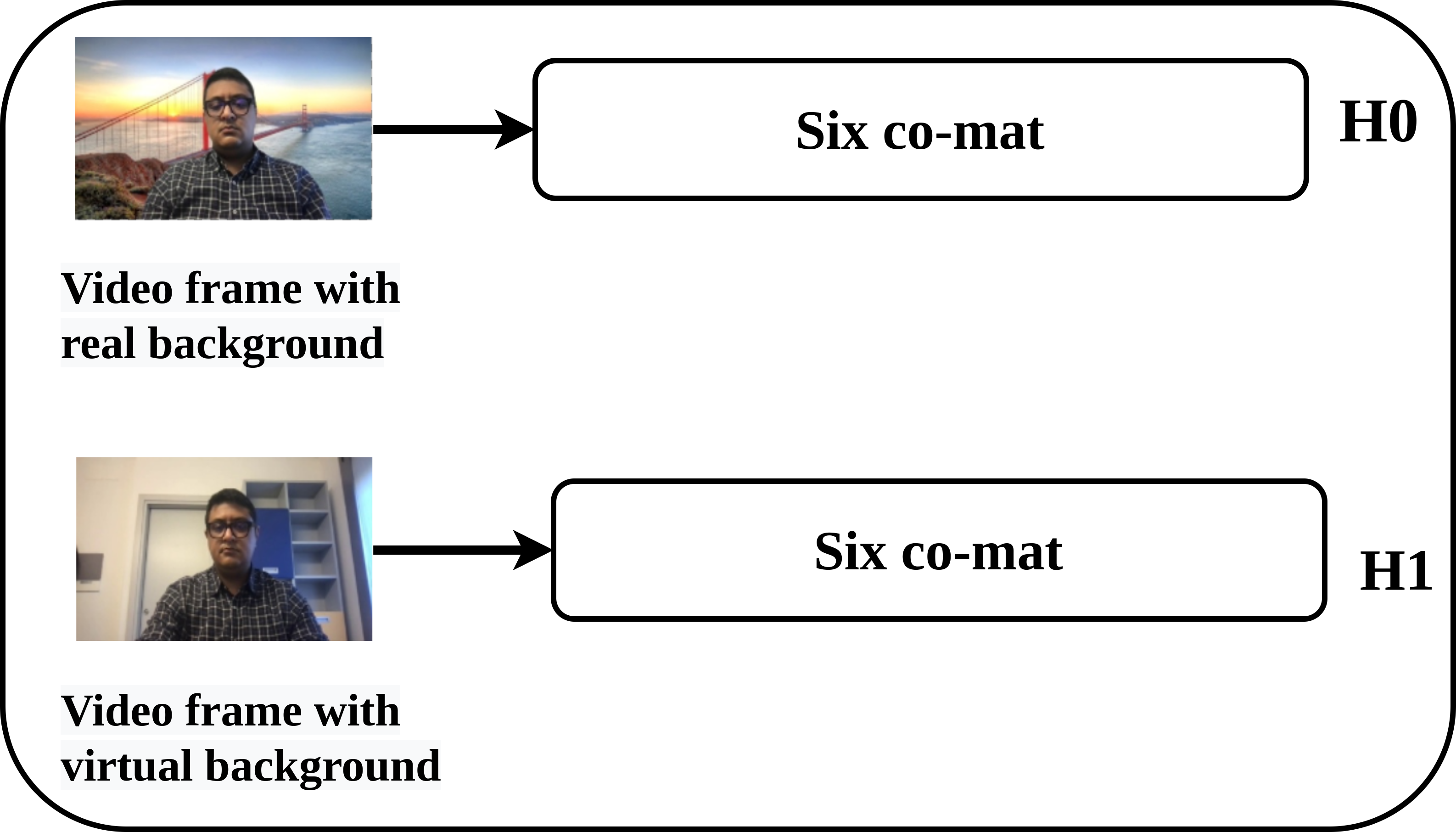} }}%
	\caption{Detection tasks for CRSPAM1372 and six co-mat features. (a) the detector is trained with CRSPAM1372 features, (b) the detector is trained with six co-mat features.}%
	\label{fig:Approach}%
\end{figure}

In our system, we consider a detection mechanism that distinguishes virtual background from real background videos by exploiting inconsistencies among spectral bands.  Specifically, we use as input to the CNN model the cross-band and spatial co-occurrence matrices~\cite{WIFS2020_CoMat}, then we train to identify the real background from videos with a virtual background. 
The problem addressed in this work is schematically depicted in Figure \ref{fig:Approach}, i.e., distinguishing videos with real background (hypothesis $H_0$, we refer as pristine) from videos with virtual one (hypothesis $H_1$, we refer as manipulate). After capturing videos from Zoom or other video conferencing software, we take advantage of the visibility of the pixels that are non-static, and appear only occasionally in video frames in the real environment when the transition between foreground and background changes (e.g., during the movement of the person in the front). In this study, we will address recent feature computation approaches CRSPAM1372 \cite{Ehsan_IWBF2018} and \textit{six co-mats} strategy \cite{WIFS2020_CoMat} that are used for forensic image detection tasks. We note that we are using only JPEG images for evaluating the robustification of the detector, since recently JPEG compression present one of the most harmful attacks in image forensics. Furthermore, two types of classifiers, SVM and CNN, are considered as a detector. Specifically, the CNN detector has only been trained in two modes: unaware and aware case. However, the SVM is trained only in an unaware modality, since the performance completely drops. Consequently, the aware case is not significant regarding the performance of the detector. In this context, the training was built according to the same scheme for the unaware detector, but without compression at the end or any attack type. Therefore, the detector is completely unaware of all attacks~\cite{nowroozi2020machine,Nowroozi2021AForensics}. In this case, we applied different attacks to the unaware detector that degraded most of the performance of an unaware version of the classifier. Thus, the detector must be aware of this attack and start training the classifier with this attack sample that we refer to as an aware detector.

\subsection{Threat Model\\}
In this work, we mainly consider an adversary that can change his real background to a virtual one during a videoconferencing call. Although the objects in the background can be identified, it is unlikely to distinguish whether the background frames are real or virtual. We assume also that the adversary can change the lightening environment and different post-processing operations such as resizing, zooming, and rotation. These conditions are usually applicable when in home environment. 
Thereby, we analyze novel attacks (we refer to physical attacks) to identify whether the considered background is real or virtual for a given videoconferencing call.

\subsection{Feature Extraction Methods\\}

In what follows, we describe the two main feature extraction approaches, namely CRSPAM1372~\cite{Ehsan_IWBF2018} and \textit{six co-ma}t~\cite{WIFS2020_CoMat}.\\

\subsubsection{The CRSPAM1372 method\\}
\label{sec.CRSPAM1372_CoMat}

In terms of the detection task, we must select a large number of features capable of capturing different types of relationships between neighboring pixels. In contrast, the feature dimensionality must be limited when training with SVM and CNN or other classifiers. Indeed, utilizing a high number of feature sets may improve modeling capabilities. However, it requires the use of multiple classifier algorithms such as ensemble classifiers ~\cite{Jessica_Ensemble2012}. In this case, the training process is challenging especially in adversary-aware modalities. In general, residual-based features have been used to identify a wide range of global manipulations \cite{Jessica_Rich2012,Jessica_SPAM2010}. The feature set is generated by evaluating the residual in all directions (e.g., $\leftarrow$, $\rightarrow$, $\uparrow$, $\downarrow$, $\nearrow$, $\searrow$, $\swarrow$, $\nwarrow$), then values truncating with a specific number of $T = 3$, and finally, co-occurrences are estimated with order $d = 2$. Many feature extraction methods were designed for the gray-scale level and cannot be applied to color images. One technique would be to extract information from the luminance channel; however, each alteration to an image changes the relationship between color channels. As a result, focusing only on the luminance channel would lose potentially relevant information. To analyze the relationships between color channels, we consider CSRMQ1, a comprehensive model for color images recently introduced in \cite{Rich_Goljan2014} for steganalysis, which is consists of two components. The Spatial Rich Model (SRMQ1) \cite{Jessica_Rich2012} and 3-D color co-occurrence are used to derive the first and second components. SRMQ1 characteristics are calculated individually for each channel to maintain the same dimensionality and then combined together. The same noise residuals SRMQ1 are considered to produce features in 3-D color components but cross the color channels. The resulting feature space has a dimensionality of 12.753 and cannot be employed with a single classifier such as SVM. Thus, for this problem, the authors of \cite{Ehsan_IWBF2018} used the SPAM686 \cite{Jessica_SPAM2010} feature set with the same method as in CSRMQ1, and they refer to CRSPAM1372. The first component in CRSPAM is derived by evaluating second-order co-occurrences of the first-order residuals $d = 2$. The features are then truncated with 3 ($T = 3$), computed separately for each channel, and then combined. Following that, in the second component, the residual co-occurrences concerning the three channels are computed. In the end, the CRSPAM feature set has 1372 dimensions in total.\\

\subsubsection{The six co-mat method\\}

It has recently been proved that generated images may be revealed by evaluating inconsistencies in pixel co-occurrences \cite{Nataraj2019DetectingGG}. The authors in \cite{WIFS2020_CoMat} compute cross-band co-occurrences (spectral) and gray-level co-occurrences (spatial), then feed to CNN to discriminate between authentic and malicious images. They claim that cross-band is more resistant to various post-processing processes, which often focus on spatial-pixel interactions rather than the features with cross-color-band. For an image $I = (\alpha, \beta, \delta)$ with a size $W \times L \times 3$, where each color channel has an offset or displacement $\Gamma = (\Gamma_\alpha, \Gamma_\beta)$ that is used with (1, 1) to compute the spatial co-occurrence matrix, and $\Gamma' = (\Gamma_\alpha', \Gamma_\beta')$ is used with (0, 0) for inter-channel. The spatial co-occurrence matrices of red, blue, and and green channels are expressed as follows:

\begin{equation}\label{eq1}
W_{\Gamma}(x,y; I_{Red})  =  \sum_{\alpha=1}^W \sum_{\beta=1}^L 
\begin{cases} 1 & \text{$if$ $\Gamma(\alpha,\beta) = x$ $and$  $I_{Red}(\alpha + \Gamma_\alpha, \beta + \Gamma_\beta) =y$}\\ 0 & \text{$otherwise$}
\end{cases}
\end{equation}

We note that similar equation~\ref{eq1} is applicable for Green and Blue channels. The CNN network's input is provided by the tensor $T_{\Gamma, \Gamma'}$, which has a dimension of $256\times 256 \times 6$ and consists of three spatial co-occurrence matrices for color channels or $[C_{\Gamma}(I_{Red}),  C_{\Gamma}(I_{Green}), C_{\Gamma}(I_{Blue})]$, as well as three cross-band co-occurrence matrices for the pairs [$I_{R-G}$], [$I_{R-B}$], and [$I_{G-B}$] or $[C_{\Gamma'}(I_{R-G}), C_{\Gamma'}(I_{R-B}), C_{\Gamma'}(I_{G-B})]$. $I_{R-B}$, $I_{G-B}$, and $I_{R-G}$, refer to the colors \textit{Red and Blue}, \textit{Green and Blue}, and \textit{Red and Green}, respectively. The construction of the cross-co-occurrence matrix (spectral) for the channels Green, Red, and Blue for an image $I = (\alpha, \beta, 1)$, where $x,y$ are integers between 0 and 255 is expressed as follows:

\begin{equation}\label{eq2}
W_{\Gamma'}(x,y; I_{R-G}) = \sum_{\alpha=1}^W \sum_{\beta=1}^L 
\begin{cases} 1  & \text{$if$ $I(\alpha,\beta,1) = x$ $and$ $I(\alpha + \Gamma_\alpha',\beta + \Gamma_\beta',2) = y$}\\ 0 & \text{otherwise}\end{cases}
\end{equation}

Similarly, the same equation~\ref{eq2} is applicable for RB and GB. Additionally, the six tensors that include the Cross-Co-Net network's ($\Gamma_{\tau, \tau'}$) input consist of the three co-occurrence matrices for the channels [$I_{Red}$, $I_{Green}$, and $I_{Blue}$] and the three cross-co-occurrence matrices for the couple [$I_{R-G}$, $I_{R-B}$, $I_{G-B}$], and defined as follows:
\begin{eqnarray}
\Gamma_{\tau, \tau'}(x,y) = [W_{\Gamma}(x,y; I_{Red}),  W_{\Gamma}(x,y; I_{Green}), \nonumber W_{\Gamma}(x,y; I_{Blue}), \\ W_{\Gamma'}(x,y; I_{R-G}), W_{\Gamma'}(x,y; I_{R-B}), \\ W_{\Gamma'}(x,y; I_{G-B})]. \nonumber
\end{eqnarray}

\section{Experimental Methodology}
\label{sec.exMethodd}
In this section, we describe our novel considered datasets~\cite{DatasetCalls} and the experimental method to assess the detector's efficiency against various test scenarios. 

\subsection{Dataset\\}
\label{sec.dataset}
\begin{figure}[h!]%
	\centering
	\subfloat[Video frames with real background]{{\includegraphics[width=14cm,height=7cm]{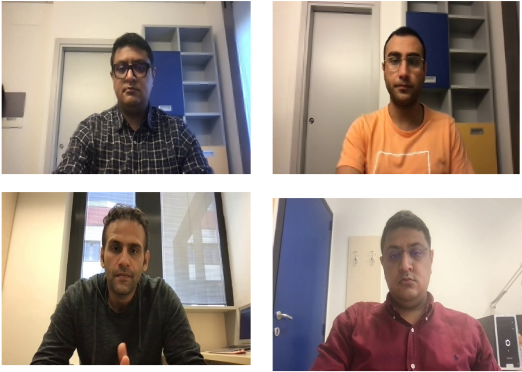} }}%
	\qquad
	\subfloat[Video frames with virtual background]{{\includegraphics[width=14cm,height=7cm]{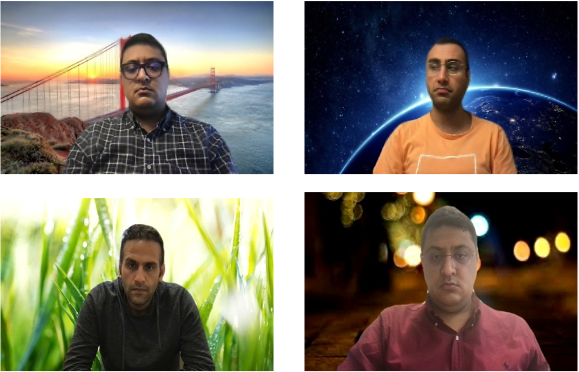} }}%
	\caption{ Examples of video frames with real and virtual background. (a) Examples of frames with real background;  (b) Examples of frames with virtual background (users considered virtual background as a real background).}%
	\label{fig:RealvsVirtual}%
\end{figure}

Due to the lack of dataset regarding the real and virtual backgrounds for videoconferencing software, we produced our own dataset and made them publicly available for the reproducibility and further applications~\cite{DatasetCalls}. In these datsets, we recorded video conferencing videos using different applications (e.g., Zoom, Google-Meet, and Microsoft Teams) and several cameras (e.g., Macbook-Pro, Apple IPAD Pro, and MSI GF65 10UX laptop) with diverse real and virtual backgrounds, changing subjects, and lighting.  The frames are then extracted from the videos and resolution of $1280\times720$. Figure \ref{fig:RealvsVirtual} shows some frame examples of videos in real backgrounds and with various virtual backgrounds. Specifically, we test our detector against several videos in a mis-match case while considering real and virtual backgrounds. We mention that the adversarial examples consist of constructing real background frames and considering them as a virtual background frames.

\subsection{Network architecture\\}
\label{sec:network}
\begin{figure}[h!]%
	\centering
	{{\includegraphics[width=\textwidth]{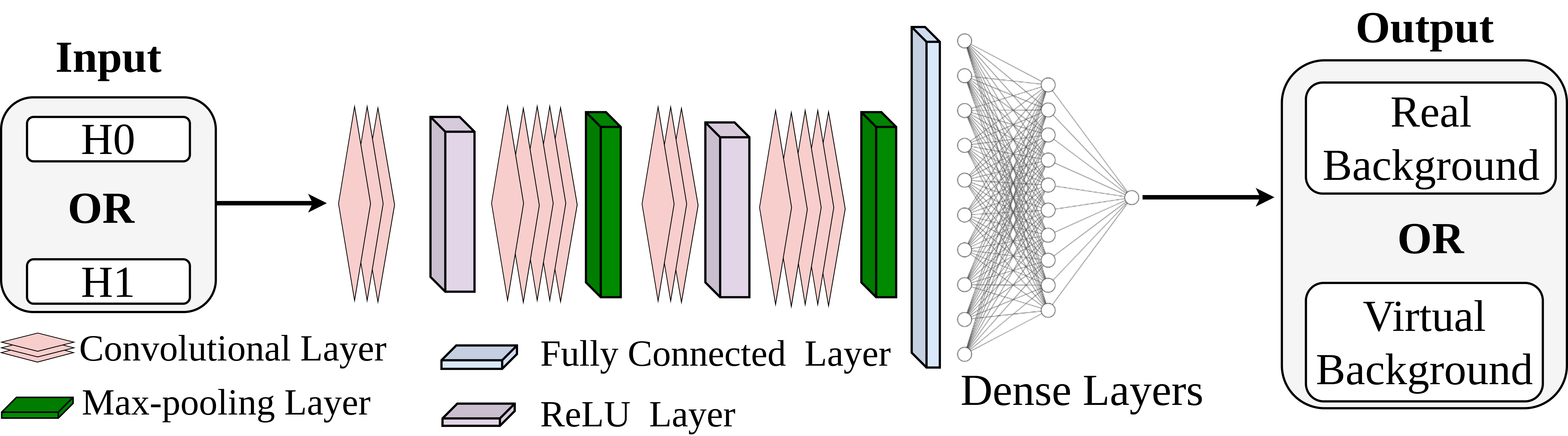} }}%
	\caption{Pipeline of the proposed network architecture. In the input, we consider H0 and H1 that are computed by the six co-mat method for a given background video frame, and the output distinguish whether the background video frame is real or virtual.}%
	\label{fig:Network}%
\end{figure}

Concerning the network architecture, we considered the network proposed in \cite{WIFS2020_CoMat}, and reduced the number of layers to fit our target application better. As illustrated in Figure.~\ref{fig:Network}, the network is composed of four convolutional layers, which are followed by two fully connected layers. A six-band input co-occurrence is present in the first input layer. In what follows, we provide a description of the network architecture:
\begin{itemize}
	\item Thirty-two filters of size $3 \times 3$ are considered for a convolutional layer with stride one and followed by a ReLu;
	
	\item Thirty-two filters of size $5 \times 5$ are considered for a convolutional layer with stride one and followed by a $3 \times 3$ max-pooling layer, plus considering 0.25 for dropout;
	
	\item Sixty-four filters of size $3 \times 3$ are considered for a convolutional layer with stride one and followed by a ReLu;
	
	\item Sixty-four filters of size $5 \times 5$ are considered for a convolutional layer with stride one and followed by a $3 \times 3$ max-pooling layer, plus considering 0.25 for dropout;
	
	\item A two dense layers by considering 256 nodes, followed by a dropout of 0.5 and a sigmoid layer. 
\end{itemize}

\subsection{Robustness Analysis\\}
\label{sec.Robus}

To produce a challenging video for detection in real scenario and to enable a low pixel drops, we captured different videos with a real and virtual background while the subject is slowly moving. Then, we applied co-mat in all the frames, separately frame by frame. We considered CRSPAM1372 for the comparison, and trained with SVM which is already proposed in the literature review. Afterward, we trained the CNN with CRSPAM1372 for an additional comparison.\\

We assess the robustness of the detectors against different types of post-processing operations. Hence, we applied different post-processing operations (see Figure \ref{fig:Approach}). Table~\ref{robustness} provides the robustness analysis of the detector that is achieved when \textit{six co-mat} matrices are considered for training the detector. Our evaluation has resulted in four different scenarios (refer Table~\ref{Summary}) for the robustness of the detector, and detailed in the following.\\

\begin{table}[h!]
\centering
\label{robustness}
\caption{Summary of the considered parameters.}
\begin{tabular}{|l|l|}
\hline
\textbf{Method}                                                                & \textbf{Considered parameters} \\ \hline
CLAHE                                                                          & 2.0, 4.0                        \\ \hline
Resizing                                                                       & 0.5, 0.8                        \\ \hline
Zooming                                                                        & 1.4, 1.9                        \\ \hline
Rotation                                                                       & 5-10                           \\ \hline
\begin{tabular}[c]{@{}l@{}}Median filtering and \\ array blurring\end{tabular} & $3 \times 3$, $5 \times 5$, $7 \times 7$ \\ \hline
Gamma correction                                                               & 0.6, 0.9, 1.3                  \\ \hline
\end{tabular}
\end{table}

\begin{figure}[th!]%
	\centering
	\subfloat[Real background in Google-Meet]{{\includegraphics[width=6.8cm,height=5cm]{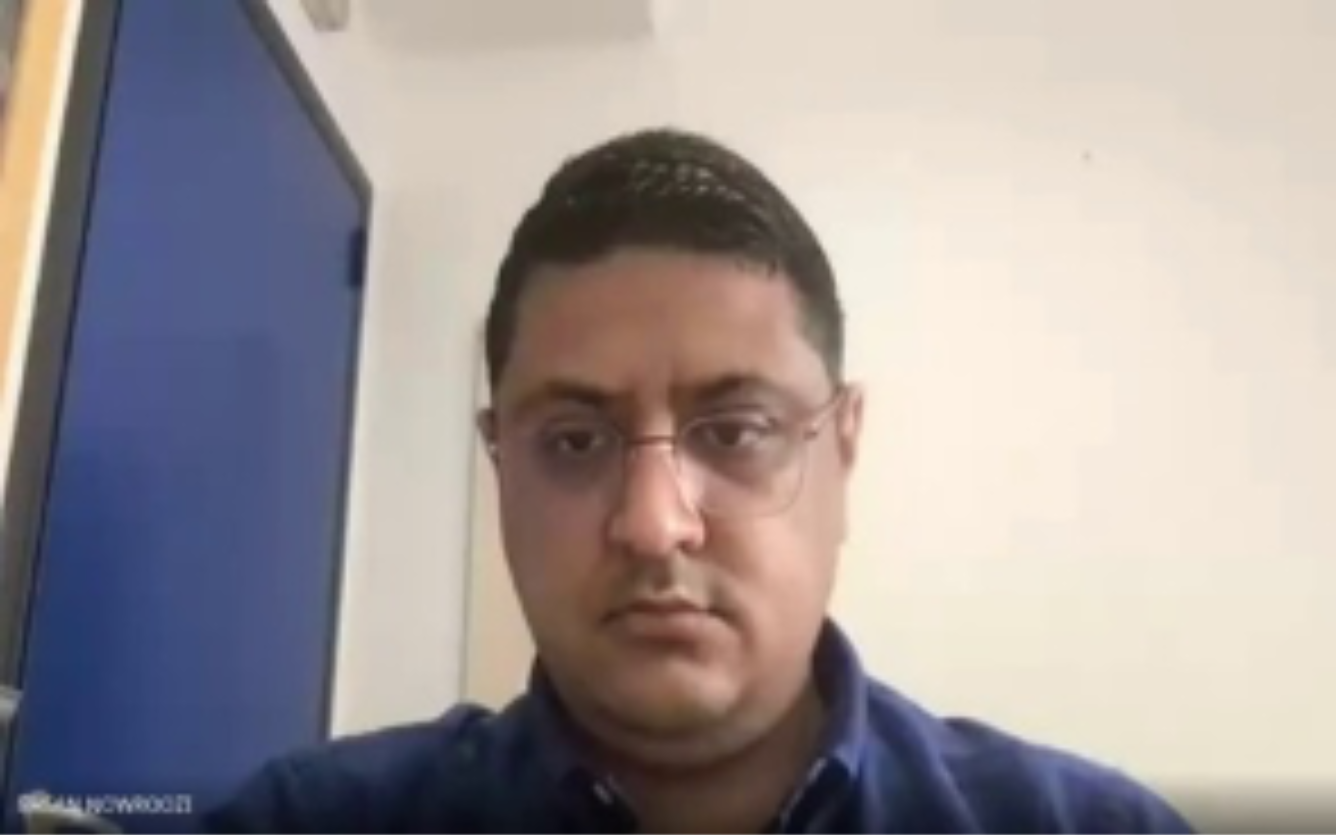}}}%
    \hspace{0.01cm}
	\subfloat[Real background in Microsoft]{{\includegraphics[width=6.8cm,height=5cm]{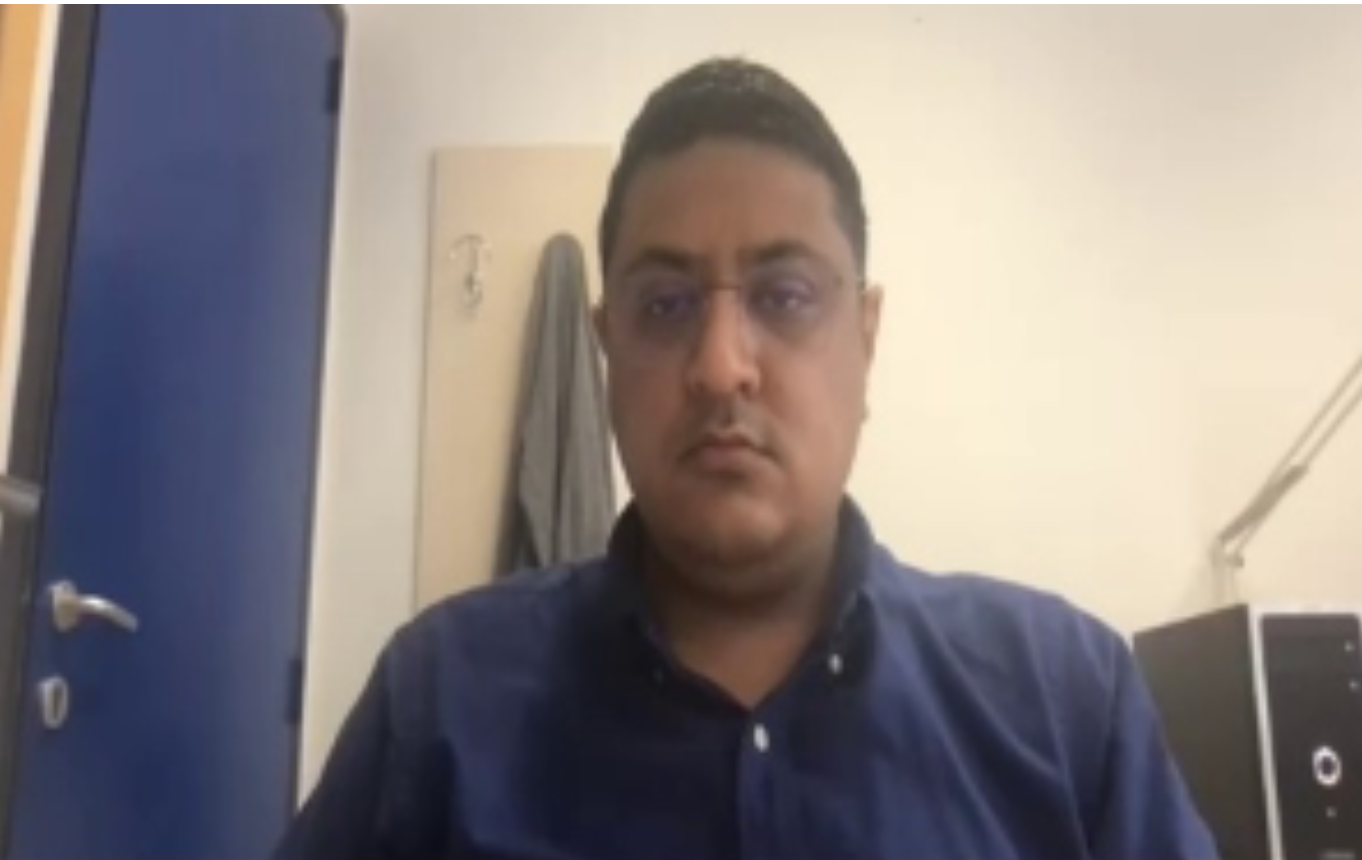}}}%
	\qquad
	\subfloat[Virtual background in Google-Meet]{{\includegraphics[width=6.8cm,height=5cm]{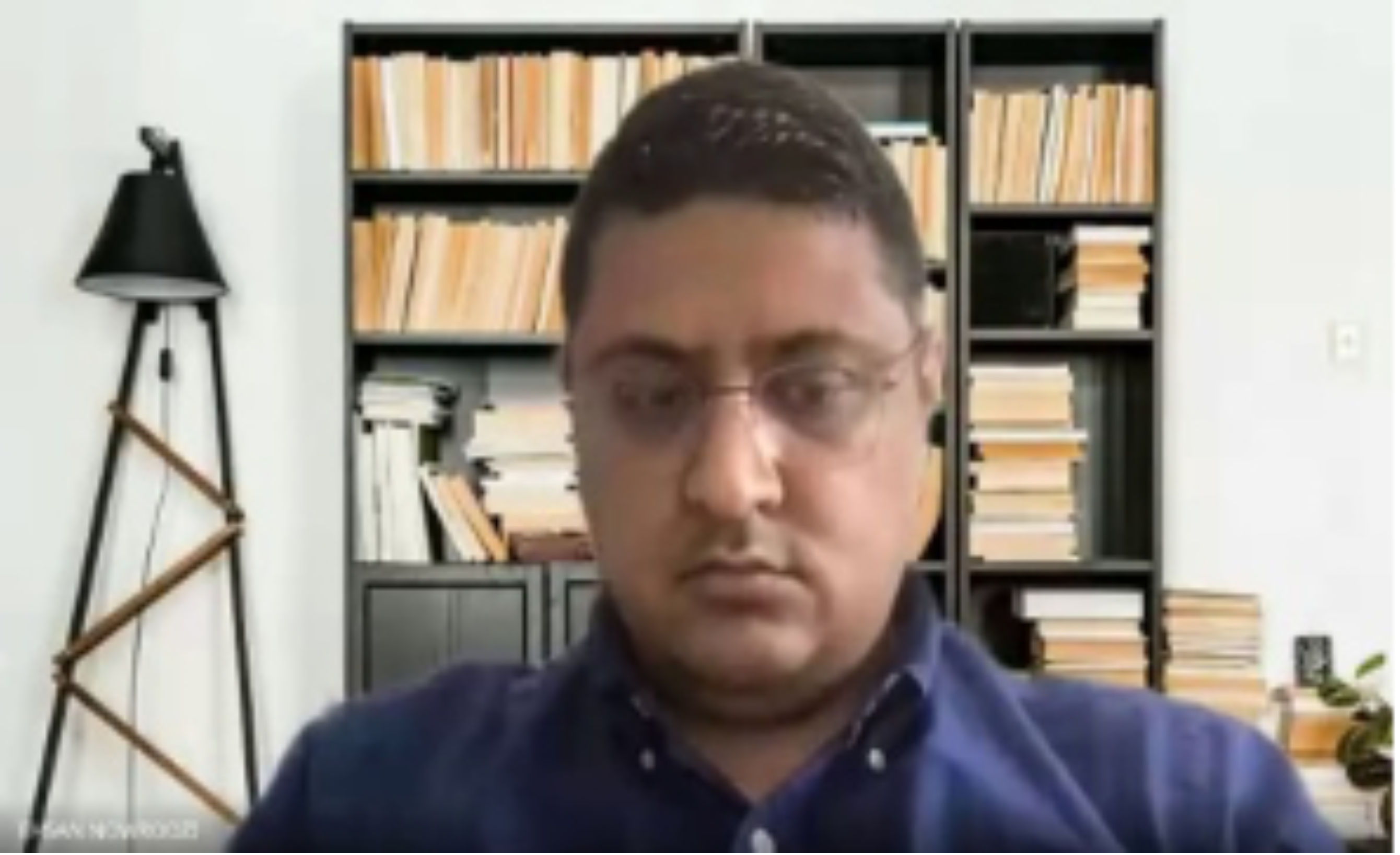} }}
	\subfloat[Virtual background in Microsoft]{{\includegraphics[width=6.8cm,height=5cm]{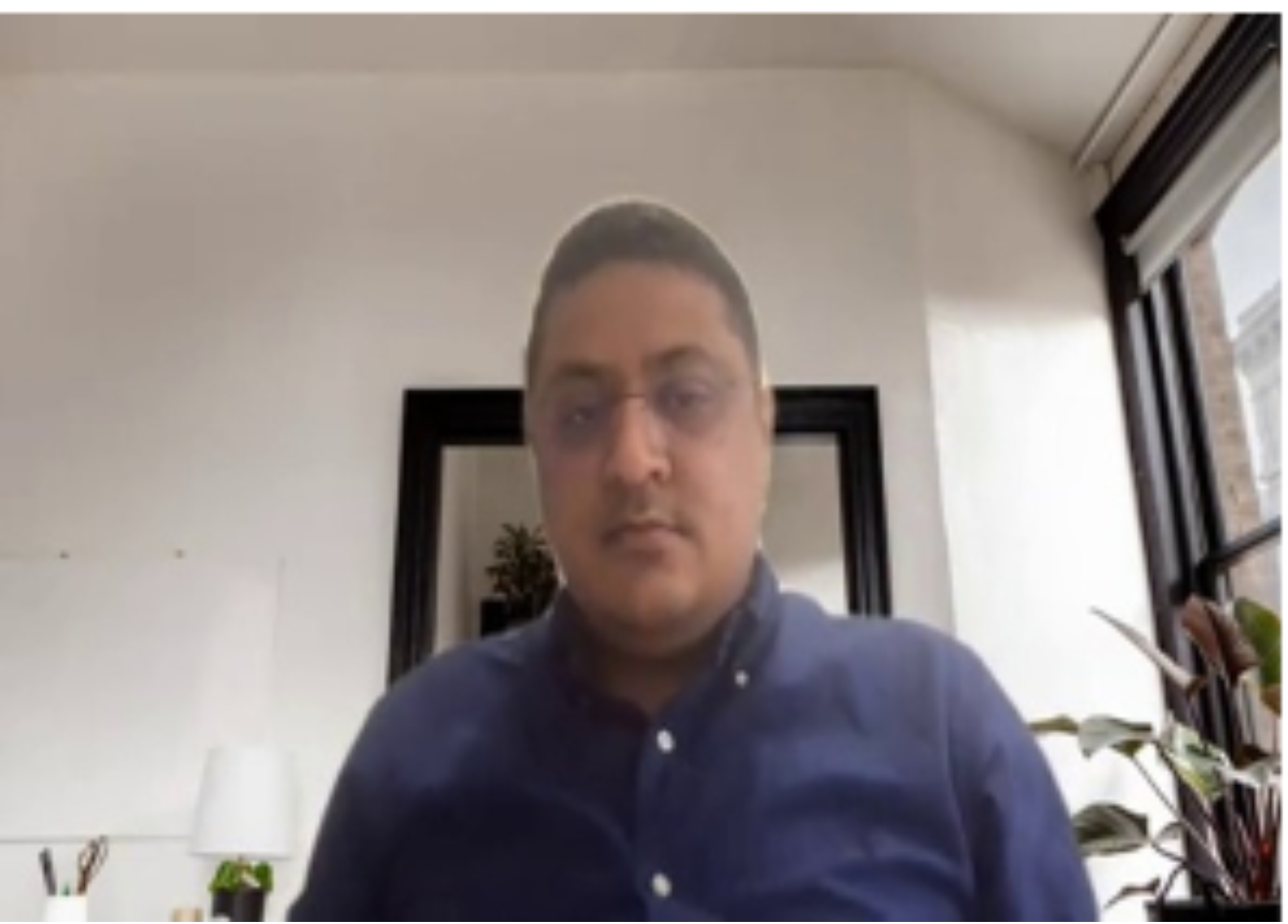}}}
	\caption{
	Examples of real and virtual background frames in Google-Meet and Microsoft Teams. (a) Real background in Google-Meet, (b) Real background in Microsoft, (c) Virtual background in Google-Meet, (d) Virtual background in Microsoft.}%
	\label{fig:RealvsVirtual_G_T}%
\end{figure}

\begin{itemize}
    \item In the first scenario, we examine the post-processing's geometric alterations, such as resizing, zooming, and rotation. We employ median filtering, average blurring for filtering operations, gamma correction, and we considered a CLAHE method for contrast modifications \cite{CLAHE}. Downscaling factors 0.8 and 0.5 are applied as resizing scaling factors, whereas upscaling values of 1.4 and 1.9 are applied with bicubic interpolation for zooming. Finally, for rotation, we applied degrees 5 and 10 with bicubic interpolation. When it comes to median filtering and average blurring, different window sizes of $3 \times 3$, $5 \times 5$, and $7 \times 7$ are considered for both filtering procedures. For gamma correction, we adjusted $\gamma$ to $\{ 0.8, 0.9, 1.2\}$, and the clip limit parameter for CLAHE was set to 2.0 and 4.0. We used Gaussian noise with standard deviations of $\{ 0.8, 2\}$ in the noise addition, and a mean of zero. Figure \ref{fig:Scheme2} shows the scheme that we considered for assessing the robustness of the detector. Furthermore, we used two post-processing techniques that were applied sequentially; hence, the robustness is compared to blurring followed by sharpening. 
    
    \item In the second scenario, the internal lighting conditions are adjusted during video capture when 75\% (low level of darkness) and 50\% (high level of darkness) of all lamps are turned on, respectively. The key reason for examining this scenario is that contrast and lighting conditions are frequently performed during counterfeiting.
    \begin{figure}[h!]%
	\centering
	{{\includegraphics[width=\textwidth]{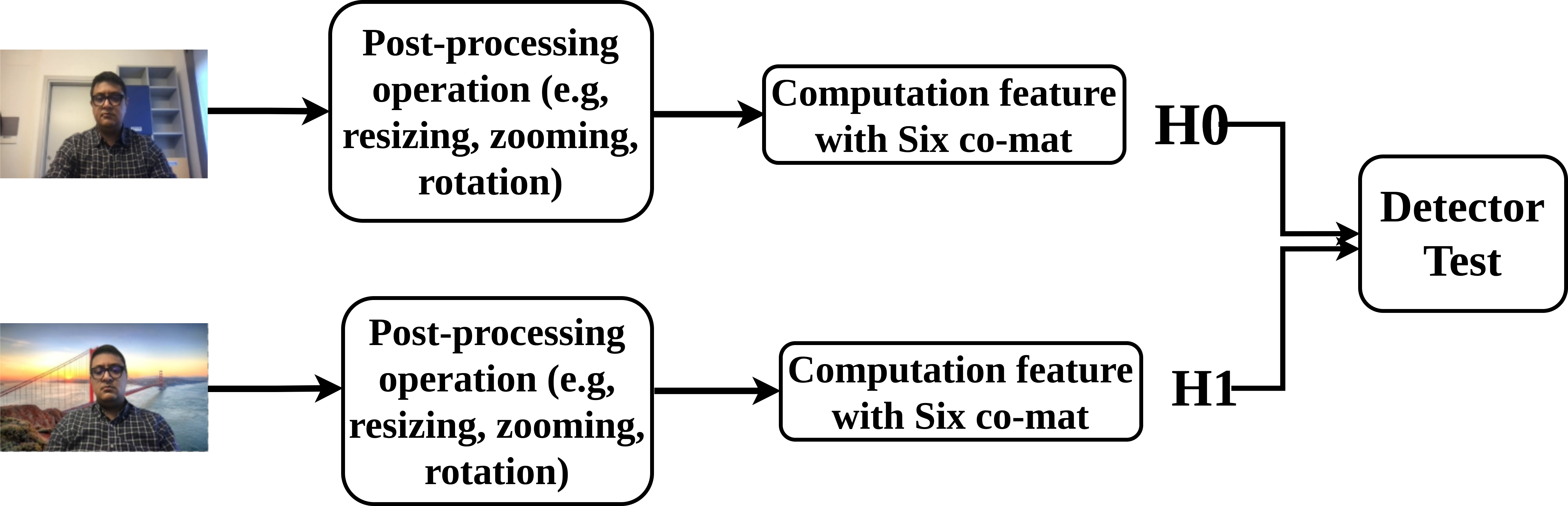} }}%
	\caption{Robustness procedure considered in this paper to evaluate the detector's performance for H0 and H1.}%
	\label{fig:Scheme2}%
\end{figure}	

    \item In the third scenario, we considered a challenging detection task to evaluate the robustness of the detector when a real background is used as virtual one. To that end, we created a harmful attack to fool the detector in an unaware case. In this situation, we suppose that an attacker has access to the subject's real background and make it a virtual one to deceive the detector. This assumption is made based on the adversary's capabilities, enabling the reconstruction of the real background frames and considering them as virtual one~\cite{Hilgefort2021SpyingCalls}. To realize this scenario, we captured a video with a real background. Then for a video with a virtual background, we consider a real background in the same place with the same lighting conditions, same camera, and a virtual background, then we start to capture the video in order to produce harmful and novel challenging attacks. Figure \ref{fig:RealvsVirtual_Attack} shows some examples when an intelligent attacker can rebuild a real background or maybe gain access to the entire background and then uses the real background as a virtual one to mislead the unaware detector.
\begin{figure}[h!]%
	\centering
	\subfloat[Real background frame in location 1]{{\includegraphics[width=6.8cm,height=5cm]{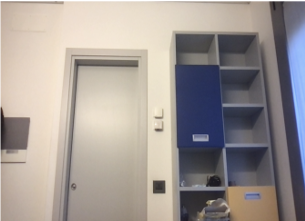} }}%
		\subfloat[Attack video frame considering real as virtual background in location 1. In this case, the real and virtual background of location 1 are the same which can be considered as a novel harmful attack.]{{\includegraphics[width=0.53\textwidth]{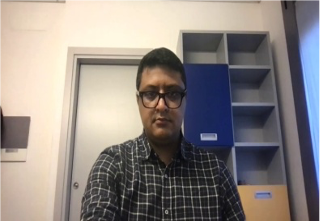} }}%
	\qquad
	\subfloat[Real background frame in location 2]{{\includegraphics[width=6.8cm,height=5cm]{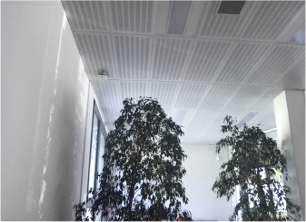} }}%
	\subfloat[Attack video frame considering real as virtual background in location 2. In this case, the real and virtual background of location 2 are the same which can be considered as a novel harmful attack.]{{\includegraphics[width=7.4cm,height=5cm]{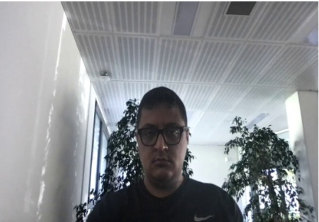} }}%
	\caption{Examples of real background and attack video frames in different locations. (a) Real background frame in location 1, (b) Attack video frame considering real as virtual background in location 1, (c) Real background frame in location 2, (d) Attack video frame considering real as virtual background in location 2.}%
	\label{fig:RealvsVirtual_Attack}%
\end{figure}	
    \item In the fourth scenario, a well-known issue with machine learning-based (ML-based) forensic tools is that they can be influenced by mismatch applications \cite{Vahid2014}, which means that the classifier performs poorly when tested with videos from a different source. As a result, we shot several videos in Google-Meet and Microsoft Teams and tested the detectors trained with Zoom videos. Figure \ref{fig:RealvsVirtual_G_T} shows some examples of Google-Meet and Microsoft Teams frames with a real and virtual background.

\end{itemize}

\begin{table}[h!]
	\centering
	\caption{Summary of the considered scenarios.}
	\label{Summary}
\begin{tabular}{|l|l|l}
\cline{1-2}
\textbf{Scenarios} & \multicolumn{1}{c|}{\textbf{Summary}}                                                                                                                                                          &  \\ \cline{1-2}
Scenario 1        & \begin{tabular}[c]{@{}l@{}}The CLAHE method with post-processing's geometric \\ alteration (e.g., median filtering, average blurring, \\ rotation, etc.).\end{tabular}                         &  \\ \cline{1-2}
Scenario 2        & \begin{tabular}[c]{@{}l@{}}Different lighting conditions, when 75\% and 50\% \\ of all lamps are turned on.\end{tabular}                                                                       &  \\ \cline{1-2}
Scenario 3        & \begin{tabular}[c]{@{}l@{}}Challenging detection task where the real background \\ is used as a virtual background.\end{tabular}                                                               &  \\ \cline{1-2}
Scenario 4        & \begin{tabular}[c]{@{}l@{}}Testing the detector against several videos captured\\  from a wide range of video-conferencing software \\ (e.g., G-Meet, Zoom, Microsoft Team, etc.)\end{tabular} &  \\ \cline{1-2}
\end{tabular}
\end{table}

We also experimented with the mismatch of the software call when videos were captured in different cameras, and we considered only Zoom videoconferencing software. In this case, we captured videos on Apple IPAD Pro and MSI GF65 10UX laptop. Table~\ref{Accuracies} provides the accuracies achieved on different model cameras in the unaware scenario.

We also tested a real-time condition (i.e., videos with real background) when we captured a video in the same lighting conditions, and the same place. Afterward, we separated the frames in a real condition and computed the co-mats. Then, we tested in the aware scenario, and we achieved 100\% accuracy if the video captured with a MacBook-Pro. Unfortunately, in real-world scenario, if the video is captured in different cameras such as Apple IPAD Pro and MSI GF65 10UX laptop, the performance of a detector in Apple IPAD decreased to 70\% but in MSI it is close to 97\%. We remark that the performance drops significantly when the videos are captured with different cameras and the only way to solve such issue is to fine tune the detector with different videos from the camera.

\begin{table}[h!]
\centering
	\caption{Accuracies on different camera model for the unaware case scenario.}
	\label{Accuracies}
\begin{tabular}{|l|l|c|c}
\cline{1-3}
\textbf{Model Camera} \hspace{3cm}                                                            & IPAD Pro & MSI GF65 10UX &  \\ \cline{1-3}
\textbf{Resolution}                                                               & 1080p    & 720p          &  \\ \cline{1-3}
\textbf{\begin{tabular}[c]{@{}l@{}}Accuracy (Test Unaware Model)\end{tabular}} & 70.45\%  & 97.63\%       &   \\ \cline{1-3}
\end{tabular}
\end{table}

\section{Analyses of Experiments}
\label{sec.ExpAna}
In this section, we discuss the analysis of our experiments by providing the considered settings, then we present the experimental results.

\subsection{Experimental setup\\}

To build our model for a detection task video with a real background from a video with a virtual one, we considered a total amount of 100000 patches for training (and validation) and 5.000 for testing per class, since the number of patches are enough for a good detection. In this case, each frame has a size with a width of 1280 and a height of 720 pixels. Then, we computed the six co-occurrence matrices from these frames. Therefore each co-mats size is $256 \times 256 \times 6$, and we applied the same split strategy for SVM when considering CRSPAM1372. We captured all the videos on MacBook-Pro with high quality and less noise.

We employed stochastic gradient descent (SGD) ~\cite{ketkar2017stochastic} as the optimizer, with a momentum of 0.9, a learning rate of 0.001, 50 training epochs, and a batch size of 20.
We built the network using the Keras API with TensorFlow as the backend~\cite{gulli2017deep}. We post-processed the frames in Python using the OpenCV package for the robustness experiments. 

Afterward, we carried out the tests on 500 frames per class from the testing set regarding each operation. To get the aware model for detecting harmful attacks (e.g., using the real background as virtual one), we considered 3600 frames for training, 700 for validation, and 450 for testing. We retrained the aware model with 50 epochs using the SGD optimizer and 0.001 as the learning rate. We used the LibSVM library package \cite{LibSVM} in the Matlab environment for training the SVM and CRSPAM1372 feature computation. We derived from each color frame 1372-dimensional features (CRSPAM1372) and fed them into the SVM classifier. Furthermore, we considered five cross-validations by adopting a Gaussian kernel to identify the kernel parameters.

\subsection{Experimental Results\\}
In what follows, we present the performance of \textit{six co-mat} and CRSPAM1372 for the previous scenarios.
\subsubsection{\textbf{Performance of Six co-mat and CRSPAM1372\\}}

\begin{figure}[h!]%
	\centering
	
	\subfloat[Red channel]{{\includegraphics[width=4.6cm,height=4cm]{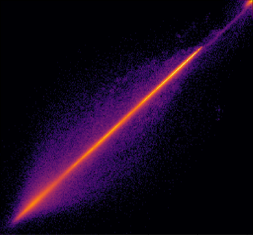} }}%
	\subfloat[Green channel]{{\includegraphics[width=4.6cm,height=4cm]{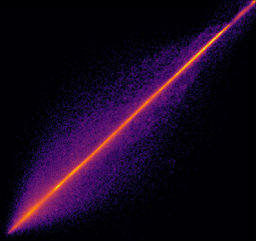} }}%
	\subfloat[Blue channel]{{\includegraphics[width=4.6cm,height=4cm]{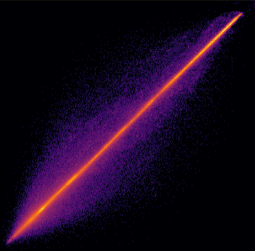} }}%
	\\
	\subfloat[Red-Green channel]{{\includegraphics[width=4.6cm,height=4cm]{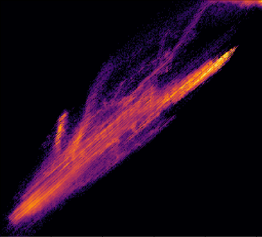} }}%
	\subfloat[Red-Blue channel]{{\includegraphics[width=4.6cm,height=4cm]{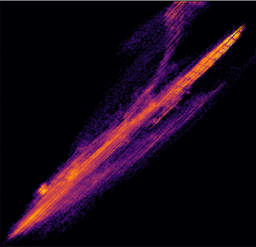} }}%
	\subfloat[Green-Blue channel]{{\includegraphics[width=4.6cm,height=4cm]{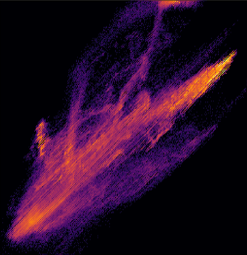} }}%
	\caption{
	Illustration of different channel co-occurrences in one real frame. The channels are: (a) Red channel, (b) Green channel, (c) Blue channel, (d) Red-Green channel, (e) Red-Blue channel, and (f) Green-Blue channel. (the size of the matrices is $250 \times 250$)}%
	\label{fig:Real_Incon}%
\end{figure}

\begin{figure}[h!]%
	\centering
	\subfloat[Red channel]{{\includegraphics[width=4.6cm,height=4cm]{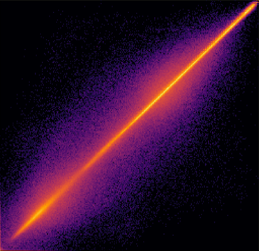} }}%
	\subfloat[Green channel]{{\includegraphics[width=4.6cm,height=4cm]{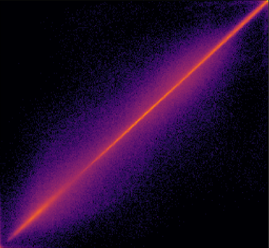} }}%
	\subfloat[Blue channel]{{\includegraphics[width=4.6cm,height=4cm]{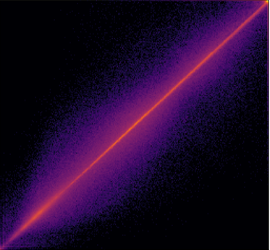} }}%
	\\
	\subfloat[Red-Green channel]{{\includegraphics[width=4.6cm,height=4cm]{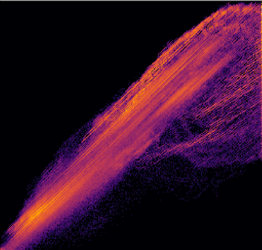} }}%
	\subfloat[Red-Blue channel]{{\includegraphics[width=4.6cm, height=4cm]{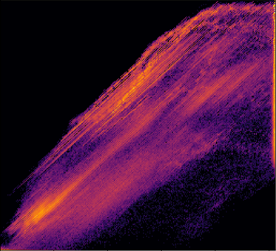} }}%
	\subfloat[Green-Blue channel]{{\includegraphics[width=4.6cm, height=4cm]{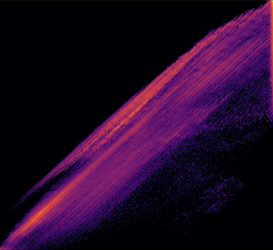} }}%
	\caption{
	Illustration of different channel co-occurrences in one virtual frame. The channels are: (a) Red channel, (b) Green channel, (c) Blue channel, (d) Red-Green channel, (e) Red-Blue channel, and (f) Green-Blue channel. (the size of the matrices is $250 \times 250$)}%
	\label{fig:Virtual_Incon}%
\end{figure}

For the \textit{six co-mat} matrices, the test set's accuracy attained is 99.80\%. However, the accuracy gained by the CRSPAM1372 is only 50.00\%, which is significantly lower than the \textit{six co-mat} method. We connected the color bands by computing separately the spatial co-occurrences on the single-color bands and the cross-band co-occurrences. Then, we demonstrated the robustness of cross-band features to standard post-processing operations (e.g., noise addition, gamma correction, median filtering,  etc.). These operations are usually focused on spatial pixel relationships instead of cross-band characteristics. By considering the matrices size $250 \times 250$ in the \textit{six co-mat} approach, we illustrate in Figure \ref{fig:Real_Incon} the inconsistencies between the spectral bands in a single real frame and spatial frame separately. For the virtual frames and with the same matrices size $250 \times 250$, we show the inconsistencies between spectral bands in Figure \ref{fig:Virtual_Incon}.  


We performed these experiments using an unaware scenario when anti-forensically-edited frames were not included in the training set. As it can be shown in the first two columns of Table~\ref{tab0}, the system is trained and tested with \textit{six co-mat} from Zoom software when the detector is unaware regarding any kinds of adversarial attacks. Additionally, we considered the aware scenario for the \textit{six co-mat} where the edited frames are including in the training set. We employed a CNN network for \textit{six co-mat}, whereas for CRSPAM1372, we considered an SVM classifier due to its simplicity. It is worth noting that we have only the the CNN and \textit{six co-mat}, because in the unaware scenario the CRSPAM1372 performance degrade to 50\%. Therefore, we considered only \textit{six co-mat} that received a good performance in the unaware case.\\

As can be shown form Table~\ref{tab0}, in the unaware scenario, the performance of the detector completely dropped to 50.00\% when we used CRSPAM1372 with a classifier SVM~\cite{Ehsan_IWBF2018}, in comparison with the CNN where we reached the performance of 83.23\%. However, in the six-co occurrence matrices when we considered CNN classifier as a detector~\cite{WIFS2020_CoMat}, we achieved 99.80\%. The key advantage of \textit{six co-mat} features over CRSPAM1372 features is greater robustness against various post-processing operations and test accuracy. The test's accuracy is reported in Table \ref{tab1} for testing the detector against different post-processings. We provide just the data for \textit{six co-mats} in this table since the test accuracy was 99.80\%, but 50.00\% for CRSPAM1372. Concerning these results, \textit{six co-mat} achieves substantially superior robustness in all post-processings, even when the operation is applied with a strong parameter. The worst-case scenario is Gaussian noise with standard deviation 2, in which the accuracy reduces to 71.60\% or somewhat less. Looking at the results of \textit{six co-mat}, we remark that the CNN network's accuracy is often around 99.00\%. The main reason for this is that performing post-processing procedures nearly completely influences the spatial relationships between pixels while having little effect on intra-channel interactions; consequently, the network may train with more robust features, resulting in a robust model.

\begin{table}[h!]
	\renewcommand\arraystretch{1.1}
	\centering
	\caption{
		Six co-mat and CRSPAM1372 accuracy.}
	\label{tab0}
\begin{tabular}{|ccc|c|l}
\cline{1-4}
\multicolumn{3}{|c|}{\textbf{Unaware  scenario}}                                                                                                                                                  & \textbf{Aware scenario}                                              &  \\ \cline{1-4}
\multicolumn{1}{|c|}{\textbf{\begin{tabular}[c]{@{}c@{}}Six co-mat \\ (CNN)\end{tabular}}} & \multicolumn{1}{c|}{\textbf{\begin{tabular}[c]{@{}c@{}}CRSPAM1372 \\ (SVM)\end{tabular}}} & \textbf{\begin{tabular}[c]{@{}c@{}}CRSPAM1372 \\ (CNN)\end{tabular}} & \textbf{\begin{tabular}[c]{@{}c@{}}Six co-mat \\ (CNN)\end{tabular}} &  \\ \cline{1-4}
\multicolumn{1}{|c|}{99.80\%}                                                              & \multicolumn{1}{c|}{50.00\%}                                                          & 83.23\%      & 99.66\%                                                              &  \\ \cline{1-4}
\end{tabular}
\end{table}
The detector's performance is efficient concerning its robustness against post-processing operations such as contrast manipulations, geometric transformations, and filtering. In this case, we separated all frames and applied different post-processing for accessing the detector's performance. We evaluate the robustness of the detector against several post-processing operations. In particular, we employed geometric transformations (resizing, rotation, and zooming), operations of filtering (blurring and median filtering), and contrast adjustment (adaptive histogram equalization which is mentioned as AHE~\cite{zuiderveld1994contrast}, and gamma correction).\\ 

For resizing operations (downscaling), we considered 0.8 and 0.5 for the scaling factor, while we applied the zooming (upscaling) with factors 1.4 and 1.9. For all the rescaling operations, we used the bicubic interpolation. Additionally, we considered angles of 5 and 10 degrees for rotation with bicubic interpolation. For the operations of filtering, we set the window size for both blurring and median filtering to 3×3, 5 × 5, and  7 × 7. For noise addition, we considered Gaussian noise with standard deviations $\gamma$ = 0.8 and 2 with zero mean. For gamma correction, we set $\gamma \epsilon$ \{0.9, 0.6, 1.3\}, while for adaptive Histogram Equalization (in particular, its refined version, Contrast Limited, implementation CLAHE~\cite{zuiderveld1994contrast}), the clip-limit parameter is set to 2.0, and 4.0 . Finally, we assessed the robustness against blurring with window size 3 × 3 followed by sharpening with kernel [[-1, -1, -1], [-1, 9, -1], [-1, -1, -1]].

\begin{table}[h!]
	\centering
	\caption{
		Detector robustness accuracy in the presence of post-processing (six co-mat (CNN)).}
	\label{tab1}
\begin{tabular}{|l|c|c|}
\hline
\textbf{Operation}                & \textbf{Parameter} & \textbf{Accuracy} \\ \hline
 \multirow{3}{*}{Median filtering} & $3 \times 3$       & 100\%             \\ \cline{2-3} 

                                  & $5 \times 5$       & 100\%             \\ \cline{2-3} 
                                  & $7 \times 7$       & 100\%             \\ \hline \hline
Gamma correction                  & 0.9                & 100\%             \\ \cline{2-3} 
                                  & 0.6                & 99.20\%           \\ \cline{2-3} 
                                  & 1.3                & 99.80\%           \\ \hline \hline
\multirow{3}{*}{Average Blurring} & $3 \times 3$       & 100\%             \\ \cline{2-3} 
                                  & $5 \times 5$       & 99.18\%           \\ \cline{2-3} 
                                  & $7 \times 7$       & 99.32\%           \\ \hline \hline
\multirow{2}{*}{CLAHE}            & 2                  & 99.80\%           \\ \cline{2-3} 
                                  & 4                  & 99.80\%           \\ \hline \hline
\multirow{2}{*}{Gaussian Noise}   & 2                  & 71.60\%           \\ \cline{2-3} 
                                  & 0.8                & 88.40\%           \\ \hline \hline
\multirow{2}{*}{Resizing}         & 0.8                & 100\%             \\ \cline{2-3} 
                                  & 0.5                & 98.20\%           \\ \hline  \hline
\multirow{2}{*}{Zooming}          & 1.4                & 99.33\%           \\ \cline{2-3} 
                                  & 1.9                & 97.30\%           \\ \hline \hline
\multirow{2}{*}{Rotation}         & 5                  & 98.25\%           \\ \cline{2-3} 
                                  & 10                 & 95.20\%           \\ \hline 
Blurring followed by sharpening   & -                  & 99.30\%           \\ \hline
\end{tabular}

\end{table}

\subsubsection{\textbf{Performance in indoor lighting conditions\\}} 

The quality of a video clip or still image is greatly influenced by indoor light. When there is too much back-light, for example, features and colors in an image are lost. As a result, distinguishing the frames of real from virtual background in the face of varying illumination conditions is a challenging task that is often applied during forgery creation \cite{Barni2018Cnn-basedPost-processing, Ehsan_IWBF2018}. It is worth noting that this kind of attacks is harmful and challenging to detect. To evaluate the detector's robustness against different lighting conditions, we captured a video when 75\% ($S_1$) and 50\% ($S_2$) of all lamps are in the physical environment. Table \ref{tab3} reports the accuracy of the detector under various lighting conditions. \\
 \begin{table}[h!]
 	\renewcommand\arraystretch{1.1}
 	\centering
 	\caption{
 		Accuracy of different lighting conditions.}
 	\label{tab3}
\begin{tabular}{l|c|c|l}
\cline{2-3}
\textbf{}                                                    & \textbf{S1} & \textbf{S2} &  \\ \cline{1-3}
\multicolumn{1}{|l|}{\textbf{Level of darkness}}             & Less dark   & More dark   &  \\ \cline{1-3}
\multicolumn{1}{|l|}{\textbf{Percentage of lamps turned on}} & 75\%        & 50\%        &  \\ \cline{1-3}
\multicolumn{1}{|l|}{\textbf{Accuracy}}                      & 100\%       & 93.66\%     &  \\ \cline{1-3}
\end{tabular}
\end{table}

 \subsubsection{\textbf{Performance while considering real background as a virtual background\\}} 
 
\begin{figure}[h!]%
 	\centering
 	\subfloat[Real background frame]{{\includegraphics[width=6.8cm,height=5cm]{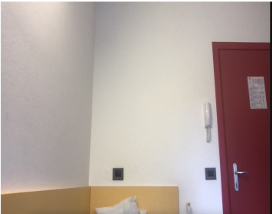} }}%
 	\subfloat[Attack background frame considering real as virtual background]{{\includegraphics[width=6.8cm, height=5cm]{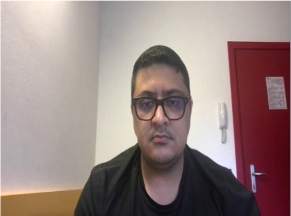} }}%
 	\caption{Real and attack background frames to test an aware model. (a) the real background frame, (b) the attack background frame considering real as a virtual background}%
 	\label{fig:RealvsVirtual_Attack1}%
 \end{figure}

In this case, we assume that an intelligent attacker can rebuild a real background or perhaps gain access to the entire background~\cite{Hilgefort2021SpyingCalls}. Therefore, the attacker can then utilize the real background as a virtual one to mislead the unaware detector. As a result, spatial relationships between pixels and intra-channel in real and video backgrounds are very close to each other; in this scenario, the attacker fully deceives the detector. Therefore, our novel strategy consists of fine-tuning the unaware detector using a variety of attack samples (we refer to the aware model). Figure \ref{fig:RealvsVirtual_Attack1} illustrates various test examples against when applied in an aware model in the real background and when the attacker considered the real background as a virtual one, and we cannot distinguish between the two backgrounds. We achieved 99.66\% training accuracy with the aware model, which was subsequently evaluated using the frames in Figure \ref{fig:RealvsVirtual_Attack1}. In this scenario, an aware model's test accuracy is 90.25\%. \\

 \subsubsection{\textbf{Performance of the detector against software mismatch\\}} 
A common problem with ML-based forensic techniques is that database mismatches might impact them; that is, the classifier performs poorly when assessed with images from the other datasets than the one used for training \cite{HigherOrder2017}. It is one of the challenging tasks in multimedia forensics. To verify that our model is efficient against software mismatch, we ran the aware model against videos shot in the same system, but using various programs such as Google-Meet and Microsoft Teams, starting with 500 Google-Meet and Microsoft Teams frames, with a size of $1072\times728$. Table \ref{tab4} shows the detector's performance when tested versus different frames taken in a different software conference call.\\ 
 \begin{table}[h!]
 	\renewcommand\arraystretch{1.1}
 	\centering
 	\caption{
 		Accuracy of aware model against mismatch applications.}
 	\label{tab4}
 	\begin{tabular}{|c|c|c|c|}
 		\hline
 		\multicolumn{1}{|c|}{\textbf{G-Meet}} & \multicolumn{1}{|c|}{\textbf{Microsoft Teams}}\\ \hline
 		\hspace{2cm} 99.80\% \hspace{2cm} & \hspace{2cm} 63.75\% \hspace{2cm} \\ \hline
 	\end{tabular}
 	
 \end{table}

As can be seen, the results obtained on Google-Meet are somewhat better than those achieved on Microsoft Teams because we discovered that Microsoft Teams saved videos in lower quality. Also, the noise has a greater impact on our detection (see Table \ref{tab1} gaussian noise results). We remark that when capturing videos in Google Meet, the storage has less noise than Microsoft Team. Additionally, we ran experiments to assess the robustness performance of the detector in the presence of noise addition and JPEG post-processing. This could be explained due to the existing harmful attacks that employs JPEG in image forensics. However, the attacker can consider other formats such as MPEG, but will result in the same behavior. Consequently, we observe that adding noise in images or frames in multimedia forensics always impacts the detector's effectiveness. Moreover, the classifier's performance drops when considering a large amount of noise. We note that adding any noise in an image or frames in multimedia forensics always affects a detector's robustness. Expectedly, if we consider a larger noise, the performance of the classifiers drops. By considering these observations, Microsoft Team has a large noise on the videos and consequently drops the detector's performance. To design such a detector that can properly operate under these conditions, we rely on an aware classifier that includes Microsoft Team videos during the training to increase its performance.

 \subsubsection{\textbf{Performance of the detector against post-processing employed before compression\\}}
 
In this scenario, we considered median filtering operation with the window size $3\times3$ before compression to assess the aware detector's performance. In the presence of processing operators that are applied before the final compression, the accuracy achieved by the model has slightly reduced when trained with \textit{six co-mat}. We note that JPEG compression is one of the most harmful laundering attacks, proven to deceive most detectors proposed. Accordingly, we considered geometric transformations (resizing) and filtering operations (median filtering) before compression. It is also noteworthy that applying different operations before compression erases the traces of the images in terms of statistical behavior and consequently its detection will be challenging. The results in Table \ref{tab5} confirm that the accuracy reduces very small in the presence of various Quality Factors (QFs). Particularly, we produced compressed version of the JPEG processed images for the case of median filtering (3 × 3), and the compression is applied with different QFs. In the following, a JPEG compressed version of the processed images for the case of resizing  (with scale factor 0.8) also was processed and compressed with different QFs.

\begin{table}[h!]
 	\renewcommand\arraystretch{1.1}
 	\centering
 	\caption{
 		Aware detector's performance in the presence of post-processing before JPEG compression.}
 	\label{tab5}
\begin{tabular}{|c|c|c|c|c|}
\hline
\textbf{QF}       & \textbf{95} & \textbf{90} & \textbf{85} & \textbf{80} \\ \hline
\textbf{Accuracy} & 97.90\%     & 94.50\%     & 94.20\%     & 94.20\%     \\ \hline
\end{tabular}
\end{table}

To clarify more, in multimedia forensics and security,  one of the major problems related to the generalization of the capabilities and other original quality videos expectedly impact the detector's performance since each setting for the quality of the video has its compression strategy. For example, the Zoom videoconferencing stores videos in MPEG4 format. Therefore, we test the detector in the same format MPEG4 and achieve 98.76\% test accuracy (video captured with the same camera). We also tested the detector when we post-processed the frames with a different compression strategy but not in the video compression, in image compression JPEG. In this case, we post-processed the frames with different quality factors and achieved good results (see Table 8 – Performance of the aware detector).
\section{Conclusion and Future Works}
 \label{sec.con}
 
Detecting whether the background in online videoconferencing software is real or virtual is an emerging research problem. In this paper, we presented a tool to determine whether a frame from a videoconferencing meeting software was produced with an artificial background. We demonstrated that our method can perfectly work as a detector to solve such problem. We proposed an experimental evaluation of the classifier that can detect virtual backgrounds in video conferencing software. Our end-to-end measurements are performed using common video conferencing tools (e.g., Google-Meet, Zoom, Microsoft Team). We considered \textit{six co-mat} matrices approach for identifying real versus virtual backgrounds in Zoom video when consumers consider a virtual background to protect their privacy. This approach exploits discrepancies among spectral color bands. Furthermore, pixel co-occurrence matrices were utilized for training the CNN network to discriminative features for virtual and real backgrounds. The results show that using \textit{six co-mat} approaches increased the detector's robustness against various test scenarios. We discovered that the CRSPAM1372 approach \cite{Ehsan_IWBF2018}, when compared to \textit{six co-mat}, is not a satisfactory approach, as we got 50.00\% accuracy. 
\subsection{Future Works}
Future studies will focus on the various scenarios: investigate the detector's performance whenever the attacker tries to modify the cross-band relationship to deceive the detector. It would be interesting to evaluate the detector's performance against various types of adversarial manipulations. Another interesting investigation related to generalization capabilities when various videos are taken with various cameras and video conferencing applications. It will be fascinating to see if the attacker uses a video rather than an image as a virtual background. The real video is captured in this example and used as a virtual background without the presence of a subject. 

\section*{Acknowledgements}

This work is funded by the University of Padua, Italy, under the STARS Grants program (Acronym and title of the project: LIGHTHOUSE: Securing the Transition Toward the Future Internet). The authors thank Professor Mauro Barni (University of Siena, Italy) for providing useful feedback to improve paper quality. Also, thanks to the students Mohammad Hajian Berenjestanaki, and Yoosef Habibi at the Department of Mathematics, the University of Padua, Italy, for helping us to create several Zoom videos \cite{DatasetCalls}.





\bibliographystyle{BST-Files/elsarticle-num}
\bibliography{main}

\end{document}